\documentclass{article}

\usepackage{arxiv}
\usepackage[T1]{fontenc}    
\usepackage{url}            
\usepackage{booktabs}       
\usepackage{amsfonts}       
\usepackage{amsmath}        
\usepackage{nicefrac}       
\usepackage{microtype}      
\usepackage{graphicx}
\usepackage{float}
\usepackage[hidelinks]{hyperref}

\title{Small Data Machine Learning: The Trilemma of Low-Sample Tabular Regimes}

\author{
  \begin{tabular}{cccc}
    \textbf{Ilya Emelyanov}\footnotemark[1] & \textbf{Dmitrii Dobrydin} & \textbf{Muhammad Hamza} & \textbf{Malak Abid Ali Khan} \\
    \footnotesize\texttt{1820262250@bit.edu.cn} & \footnotesize\texttt{1820262226@bit.edu.cn} & \footnotesize\texttt{7520250266@bit.edu.cn} & \footnotesize\texttt{malak@bit.edu.cn}
  \end{tabular} \\[2ex]
  \textbf{Beijing Institute of Technology, Beijing, China}
}

\begin{document}
\maketitle
\footnotetext[1]{Corresponding author.}

\begin{abstract}
Small tabular datasets leave few observations for choosing a model and assessing its predictions. We examine these competing demands in controlled simulations and on six public datasets, using SmallGBM and SmallMLP as fixed-configuration case studies. Three results illustrate why the choice of evaluation matters. First, at $n=100$, five-fold cross-validation overstated the gain from tuning XGBoost by 0.063 ROC-AUC on random labels, while tuning improved independent-test ROC-AUC by 0.0078 on a Gaussian mixture. Second, when 20\% of binary training labels were flipped, the median within-trial ROC-AUC loss was 0.0210 for SmallGBM and 0.0778 for a fixed 200-tree XGBoost configuration. A separate regression experiment showed the sensitivity of squared-error training to heavy-tailed responses. Third, in a heteroscedastic regression experiment, locally weighted SmallMLP intervals covered about one percentage point more test responses than ordinary split conformal intervals at both tested dimensions. Their mean widths also increased. On a Gaussian classification task with linear log odds, logistic regression achieved the highest mean ROC-AUC at all seven training sizes. These experiments identify settings in which a small-data model helps and quantify the cost of its assumptions.
\end{abstract}

\section{Introduction}

When observations are expensive to acquire, a practitioner must decide how much data to spend on fitting, model selection and assessment. We use the term \emph{trilemma} as an organizing framework for three objectives: limiting estimation variance, controlling approximation bias and quantifying uncertainty. The appropriate trade-off depends on the data-generating process and on the metric used to compare predictions.

Boosted trees and regularized neural networks are established methods for tabular prediction \cite{chen2016xgboost,ke2017lightgbm,kadra2021well,grinsztajn2022tree}. With a small validation sample, searching a set of configurations can favor those that happened to score well on that sample \cite{cawley2010selection,varoquaux2018cross}. This makes two questions worth distinguishing. How well does validation estimate the gain from tuning? How much does the configuration selected by that search improve predictions after refitting on all available training data?

We examine these questions using two fixed-configuration estimators. SmallGBM combines randomized split selection with robust aggregation of classification residuals. SmallMLP uses dataset size and dimensionality to set its architecture and regularization. A third question arises when using its predictions to construct intervals: how do interval width, overall coverage and coverage in a high-noise subgroup change when calibration residuals receive local weights? We compare the models on paired synthetic trials and on six public classification and regression datasets. The controlled trials vary sample size, target contamination and noise scale, while the dataset comparisons measure performance under specified splitting protocols.

\subsection*{Related work}

Several approaches to tabular prediction make model complexity a choice during training. TabNet uses sequential attention to select features for each observation \cite{arik2021tabnet}. A comparison of residual networks, tabular Transformers and boosted trees found that their relative performance varied across tasks under a common evaluation protocol \cite{gorishniy2021revisiting}. Automated systems can also combine models. Auto-sklearn uses prior evaluations to initialize a search over pipelines \cite{feurer2015automl}, whereas AutoGluon-Tabular combines predictors through multilayer stacking \cite{erickson2020autogluon}. Our controlled search experiment examines a single part of this process, the gain estimated when selecting among configurations of one model family.

Uncertainty estimates can come from a Gaussian process posterior \cite{rasmussen2006gpml} or an ensemble of independently trained neural networks \cite{lakshminarayanan2017ensembles}. Conformalized quantile regression calibrates fitted conditional quantiles and can vary interval width with the input \cite{romano2019cqr}. Our interval experiment instead holds the fitted regression function fixed and varies how calibration residuals contribute to the interval radius.

The relevant sampling unit depends on the scientific question. Materials informatics draws on both measured and computed properties \cite{ramprasad2017materials}, and particle-physics analyses use detector measurements for tasks such as separating signal events from background \cite{radovic2018physics}. In the public-data experiments below, we specify whether a held-out prediction concerns a new observation, a new patient or a new physical design.

\section{Fundamental Bottlenecks and Mathematical Principles}
\label{sec:theory}

We describe the loss derivatives, validation estimates and prediction intervals used in the experiments. Each result below states its sampling assumptions.

\subsection{Vulnerability to Outliers and Robust Estimation}
For a fixed leaf with sample index set $I$, the regularized second-order update for XGBoost with $L_2$ leaf penalty $\lambda$ is \cite{chen2016xgboost}
\begin{equation}
\label{eq:taylor_leaf}
w^* = -\frac{\sum_{i \in I} g_i}{\sum_{i \in I} h_i + \lambda}
\end{equation}
where $g_i$ and $h_i$ are the first and second derivatives with respect to the current score. For squared-error regression, $g_i=\widehat y_i-y_i$ and $h_i=1$, so a single arbitrarily large response can move the update arbitrarily far. For binary logistic loss, $g_i=p_i-y_i$ lies in $[-1,1]$ and $h_i=p_i(1-p_i)$. Flipping a binary label changes its gradient by one in magnitude at a fixed prediction. Effects on later splits and scores depend on the fitted ensemble.

SmallGBM 1.5.0 fits classification pseudo-residuals $r_i=y_i-p_i$. Let $m_I$ be their median within a nonempty node and $s_0>0$ the prior scale (default $0.5$). The native implementation computes the node center $c_I$ as
\begin{equation}\label{eq:median_leaf}
c_I=\begin{cases}
m_I,& |I|\leq30,\\
\sum_{i\in I} a_i r_i / \sum_{i\in I} a_i,& |I|>30,
\end{cases}
\qquad a_i=(|r_i-m_I|+s_0)^{-1}.
\end{equation}
For small nodes, the scale $s_I$ is $1.4826$ times the median absolute deviation about $m_I$, with $s_I=s_0$ when that deviation is zero. For larger nodes it is the weighted standard deviation about $c_I$ using the same weights $a_i$. The update shrinks the center toward the parent value $w_P$:
\begin{equation}
\label{eq:mad_shrinkage}
u_I=\frac{s_I^2}{|I|s_0^2},\qquad
\gamma_I=\frac{u_I^{3/2}}{1+u_I^{3/2}},\qquad
w_I=(1-\gamma_I)c_I+\gamma_I w_P.
\end{equation}
The finite-sample resistance of a median to extreme observations concerns a fixed node \cite{huber2009robust}. An ensemble also learns its node boundaries and re-estimates residuals after each tree. Section~\ref{sec:heavy_tailed_noise} evaluates the complete classifiers under label flips.

\subsection{Validation Uncertainty and Adaptive Complexity}
Consider a finite set $\Lambda$ of predictors chosen independently of a hold-out sample. If the validation observations are independent and identically distributed and the per-observation loss lies in $[0,1]$, Hoeffding's inequality and a union bound give, with probability at least $1-\delta$,
\begin{equation}
\label{eq:val_bound}
\sup_{\lambda\in\Lambda}|\widehat R_{\rm val}(\lambda)-R(\lambda)|
\leq \sqrt{\frac{\log(2|\Lambda|/\delta)}{2n_{\rm val}}}.
\end{equation}
This bound describes a single independent validation sample. ROC-AUC compares pairs of observations, and the training folds in cross-validation overlap. We therefore estimate selection optimism in Section~\ref{sec:tuning_trap} using independent test data and matched training sizes.

SmallMLP sets the classification width from the training size $n$, feature count $d$ and class count $K$. Write $b_2=2$ and $b_K=2K$ for $K>2$. The first hidden layer has width
\begin{equation}
\label{eq:width_w1}
w_1 = \max\left(b_K, \; \min\left(\left\lfloor \alpha \sqrt{n K \sqrt{d}} \right\rfloor, \; 4n\right)\right)
\end{equation}
with coefficient $\alpha=6$. Subsequent classification layers use $\beta=0.7$ and
\begin{equation}
\label{eq:width_wl}
w_l = \max\left(K, \; \left\lfloor w_1 \cdot \beta^{l-1} \right\rfloor\right)
\end{equation}
For regression, all hidden layers have the same width $\min(128,\max(16,2n))$. Both estimators set dropout by
\begin{equation}
\label{eq:dropout}
p_{\text{drop}} = \min\left(\max\left(3 \cdot \frac{d}{n}, \; 0.1\right), \; 0.5\right)
\end{equation}
These choices fix the architecture before training. SmallMLP still reserves an internal validation subset for early stopping.

\subsection{Marginal Coverage and Locally Weighted Prediction Intervals}
\label{sec:uq}

Let $\widehat f$ be a regression predictor fitted without using the calibration or test observations. For $m$ calibration observations, sort the absolute residuals $R_i=|Y_i-\widehat f(X_i)|$ as $R_{(1)}\le\cdots\le R_{(m)}$. Ordinary split conformal uses the same half-width $R_{(k)}$ for every test input, where $k=\lceil(m+1)(1-\alpha)\rceil$ and $R_{(m+1)}=+\infty$. If the $m$ calibration observations and the new observation are exchangeable conditional on the fitted predictor, then
\begin{equation}
\label{eq:conformal_cov}
\mathbb P\!\left\{Y_{\rm new}\in C_{\rm split}(X_{\rm new})\right\}\geq 1-\alpha,
\qquad C_{\rm split}(x)=[\widehat f(x)-R_{(k)},\,\widehat f(x)+R_{(k)}].
\end{equation}
When the scores are almost surely distinct, the new score has uniform rank among the $m+1$ scores, giving marginal coverage $k/(m+1)$ \cite{vovk2005algorithmic, angelopoulos2023gentle}. This probability averages over the calibration set and the new observation. Coverage within a particular noise regime can differ.

The experimental local procedure uses embeddings $h(x)$ obtained with the same feature standardization and trained backbone as the regression predictor. Its Gaussian kernel uses the median pairwise calibration distance as its bandwidth,
\begin{equation}
\label{eq:h_cal}
h_{\text{cal}}=\operatorname{median}_{1\leq i<j\leq m}\|h(X_i)-h(X_j)\|_2 .
\end{equation}
For test input $x$, the calibration weights are $\exp[-\|h(x)-h(X_i)\|_2^2/(2h_{\rm cal}^2)]$. The procedure assigns unit weight to a test score at $+\infty$ and takes the weighted $(1-\alpha)$ quantile. If the bandwidth is effectively zero it uses one. When the quantile is infinite, it caps the radius at the largest calibration residual. The test-dependent weights and the cap preclude transferring the unweighted rank argument in Equation~\eqref{eq:conformal_cov}. Other localized methods derive guarantees by calibrating the quantile level \cite{guan2023lcp} or randomizing the localization step \cite{hore2025rlcp}. We evaluate this particular procedure empirically.

\section{Real-World Applications and Experimental Methodology}

\subsection{Real-World Benchmark Domains}
The public classification tasks cover clinical outcomes (Hepatitis), diagnostic measurements (Wisconsin Diagnostic Breast Cancer), voice recordings (Parkinson's) and radar returns (Ionosphere). The regression tasks measure yacht resistance and simulated building energy demand. Repeated measurements from one person, hull design or building geometry can share predictive information. We specify the prediction unit and grouping rule for each dataset in Section~\ref{sec:real_results}.

\subsection{Comparators and Implementations}
SmallGBM 1.5.0 uses the aggregation rule in Section~\ref{sec:theory} and randomized split candidates. SmallMLP 0.4.0 uses two hidden layers by default, weight decay $1/n$ and internal early stopping. Throughout, a fixed configuration means that no hyperparameter search used that benchmark's test scores. The five-fold training-only search in Section~\ref{sec:tuning_trap} is explicitly identified. The regression interval experiment uses a separate calibration set.

XGBoost provides tree comparators with the loss and configuration specified for each experiment. A linear baseline tests the contribution of nonlinear structure. The interval experiment also includes a Gaussian process with a fitted noise term. As context, TabPFN v2 uses pretrained in-context prediction on small and medium-sized tables \cite{hollmann2025tabpfn}. Our direct comparisons below concern the models plotted or tabulated.

\begin{table}[htbp]
\centering
\caption{What each comparison measures.}\label{tab:baselines}
\small
\begin{tabular}{@{}p{.28\linewidth}p{.63\linewidth}@{}}\toprule
Comparison & Quantity estimated \\\midrule
Tuned and default XGBoost & Matched-size validation optimism and independent-test gain after refitting \\
Gaussian classification & Test ROC-AUC, accuracy and log loss for linear and tree models over seven training sizes \\
Target corruption & Within-trial change in test performance as training targets are perturbed \\
Prediction intervals & Test coverage, high-noise subgroup coverage and mean interval width \\
Public datasets & Performance of fixed configurations on held-out folds under each dataset's splitting rule \\
\bottomrule\end{tabular}
\end{table}

\subsection{Development and Evaluation}\label{sec:meta_protocol}
\begin{figure}[H]
\centering
\includegraphics[width=0.68\textwidth]{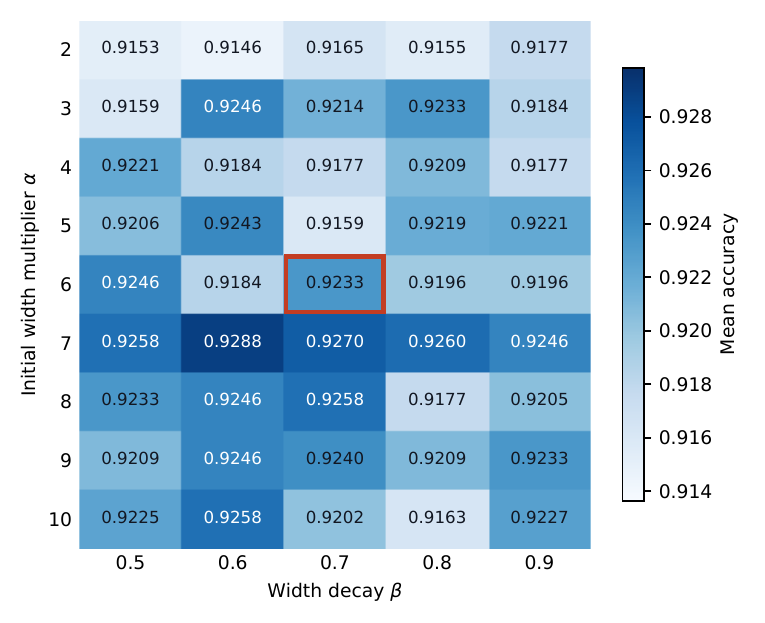}
\caption{Mean accuracy on Iris, Wine and Statlog Heart, averaged equally across datasets. The outlined cell marks the coefficients used in the benchmarks ($\alpha=6$, $\beta=0.7$).}
\label{fig:meta_surface}
\end{figure}

The classification width coefficients and the leaf-size threshold were fixed for the reported evaluations. Figure~\ref{fig:meta_surface} examines a development grid over network widths. The subsequent public-data comparison uses held-out predictions within each dataset. Prior use of a public dataset during model development can still affect dataset-level comparisons. We therefore treat the public results as benchmark estimates and document known prior exposure.

\section{Empirical Pathologies and Performance Boundaries in the Low-Sample Regime}

\subsection{Validation and Learning Curves}

\subsubsection{Tuning gain and cross-validation optimism at \texorpdfstring{\(n=100\)}{n=100}}
\label{sec:tuning_trap}

We compared three classifiers on a Gaussian mixture, a nonlinear interaction task, and a null task with labels independent of the features. For each task and each of 20 seeds, all three methods received the same 100 training observations with 10 features. We scored the fitted models on a separate sample of 5,000 observations drawn from the same distribution.

The first two methods used XGBoost 3.4.1. The baseline used the library's default model hyperparameters. For the tuned model, we evaluated 25 candidates: the default configuration and the 24 combinations of \texttt{max\_depth} \(\in\{2,3,6\}\), \texttt{learning\_rate} \(\in\{0.03,0.1\}\), \texttt{n\_estimators} \(\in\{50,200\}\), and \texttt{min\_child\_weight} \(\in\{1,5\}\). All candidates used the same five shuffled, stratified folds (80 training and 20 validation observations per fold). We selected the configuration with the highest mean validation ROC-AUC, with the default listed first to break ties in its favor, and refitted it on all 100 observations. We also fitted default XGBoost and SmallGBM 1.5.0 on all 100 observations, using SmallGBM's default model hyperparameters.

\begin{figure}[H]
    \centering
    \includegraphics[width=0.98\textwidth]{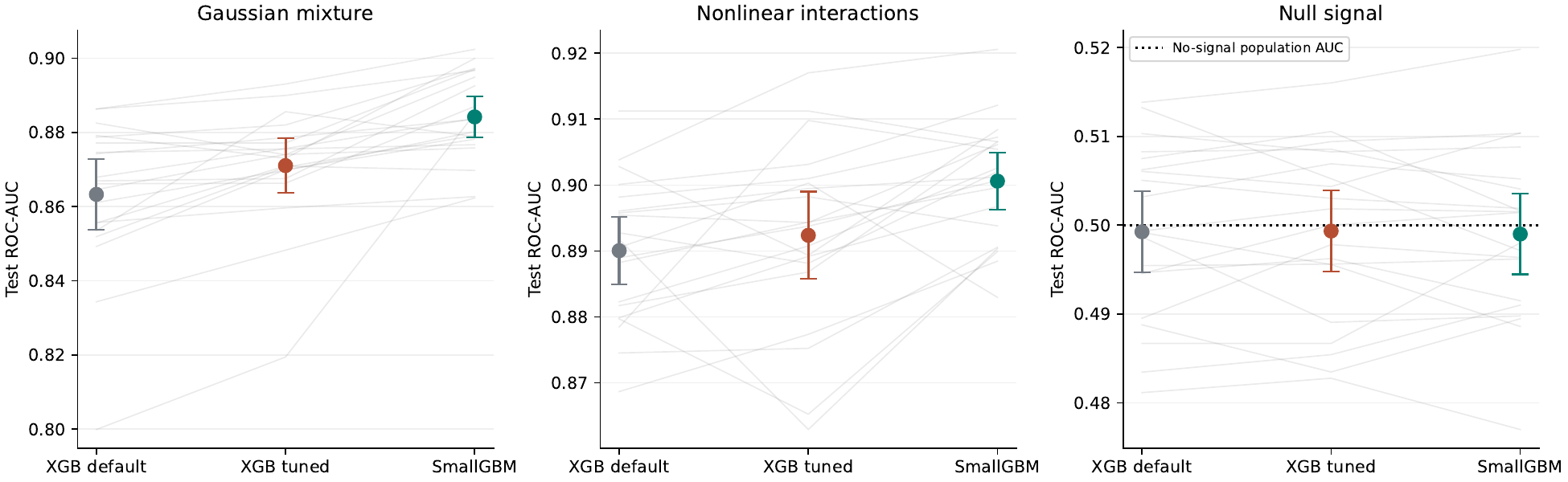}
    \caption{Test ROC-AUC after training on 100 observations. Each grey line connects the three methods on the same trial. Points show means across 20 seeds and error bars show pointwise 95\% \(t\)-intervals.}
    \label{fig:tuning_comparison}
\end{figure}

On the Gaussian task, tuning improved mean test ROC-AUC from 0.863 to 0.871 (paired gain \(+0.0078\), 95\% CI \([+0.0031,+0.0124]\)). On the nonlinear task the corresponding scores were 0.890 and 0.892 (paired gain \(+0.0023\), 95\% CI \([-0.0033,+0.0080]\)). SmallGBM had the highest mean test ROC-AUC on these two tasks, at 0.884 and 0.901, respectively. On the null task all three means were approximately 0.499, close to the population ROC-AUC of 0.5 (Figure~\ref{fig:tuning_comparison}).

\begin{figure}[H]
    \centering
    \includegraphics[width=0.92\textwidth]{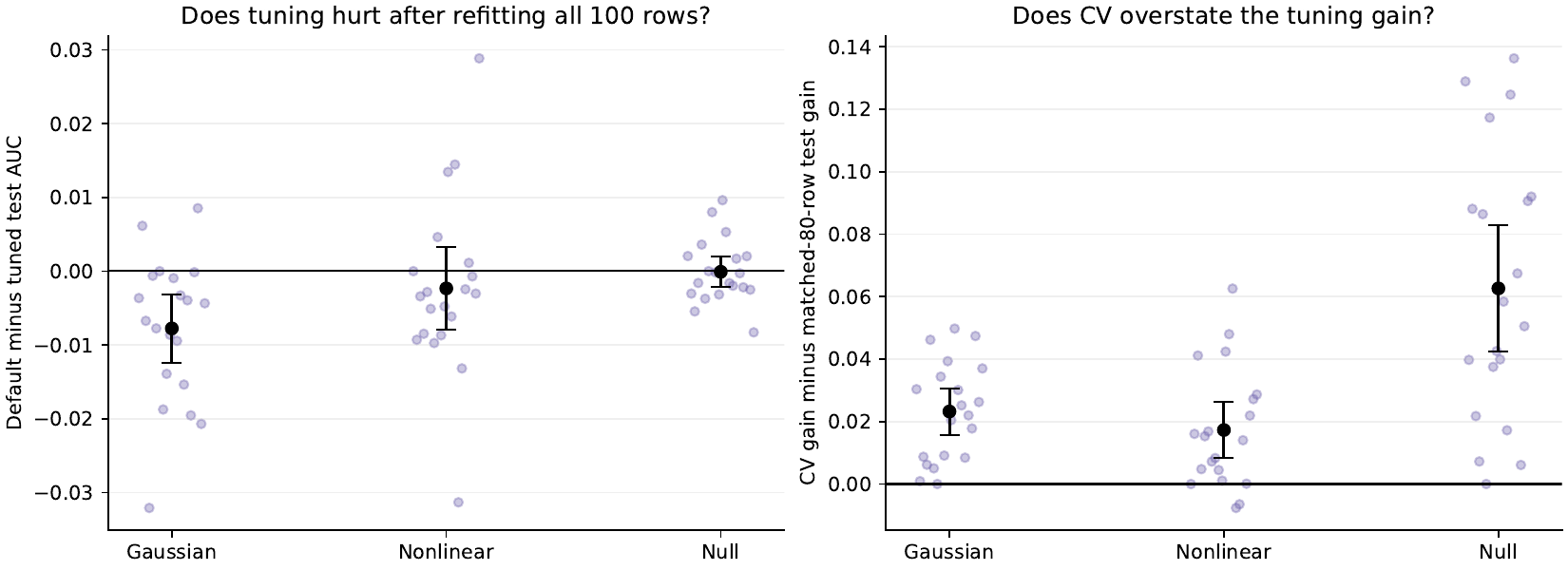}
    \caption{Paired ROC-AUC differences across 20 seeds. The left panel shows default minus tuned test ROC-AUC after fitting on all 100 observations. Negative values favor tuning. The right panel shows how much the validation gain exceeded the independent-test gain when both were measured using models trained on 80 observations. Dots show individual trials and black markers show means with pointwise 95\% \(t\)-intervals.}
    \label{fig:cv_optimism}
\end{figure}

To measure how well cross-validation estimated the benefit of tuning, we computed the selected configuration's mean validation ROC-AUC minus the default configuration's on the same folds. For comparison, we refitted both configurations on each original 80-observation training fold, scored them on the independent test set, and averaged their paired test differences over the five folds. Subtracting this matched-size test gain from the validation gain gives the quantity in the right panel of Figure~\ref{fig:cv_optimism}.

Cross-validation overstated the mean gain by 0.023 ROC-AUC on the Gaussian task, 0.017 on the nonlinear task, and 0.063 on the null task. On the null task, the apparent validation gain was 0.062 while the matched-size test gain was \(-0.0006\). Thus the search identified a validation advantage even when the features carried no information about the labels. On the Gaussian task, the apparent advantage was also optimistic, although tuning improved the final model's test score. The experiment measures selection optimism for one grid and one training size. Testing the scaling in Equation~\eqref{eq:val_bound} would require varying those quantities.

\subsubsection{Sample size and ROC-AUC on a Gaussian mixture}
\label{sec:n_crossover}

We drew equally likely binary labels and generated ten features from two Gaussian classes:
\[
\begin{gathered}
Y\sim\operatorname{Bernoulli}(1/2),\qquad
X\mid Y=y\sim\mathcal{N}\bigl((2y-1)\mu,I_{10}\bigr),\\
\mu=v/\|v\|_2,\qquad v=(1,0.75,-0.5,0,0,0,0,0,0,0)^\top .
\end{gathered}
\]
The Bayes score $S=\mu^\top X$ has distribution $\mathcal{N}(1,1)$ in class 1 and $\mathcal{N}(-1,1)$ in class 0. Thresholding at zero gives population accuracy $\Phi(1)=0.8413$. For independent draws from the two classes, their score difference has distribution $\mathcal{N}(2,2)$, so the population ROC-AUC is $\Phi(\sqrt{2})=0.9214$, where $\Phi$ is the standard normal distribution function.

For each of 20 seeds, all models shared a test set of 10,000 independent observations and nested training sets of 100, 200, 500, 1,000, 2,000, 5,000 and 10,000 observations. SmallGBM used its default configuration. Both XGBoost configurations used histogram trees of depth 3. \texttt{XGBoost-50} used 50 trees, a learning rate of 0.10, and half the features per tree. \texttt{XGBoost-200} used 200 trees, a learning rate of 0.05, and all features per tree. Because the population log odds are $2\mu^\top X$, we added logistic regression as a linear control when revisiting this experiment. It used $L_2$ regularization ($C=1$), an intercept, the L-BFGS solver and the same unscaled features and splits.

\begin{figure}[H]
    \centering
    \includegraphics[width=0.98\textwidth]{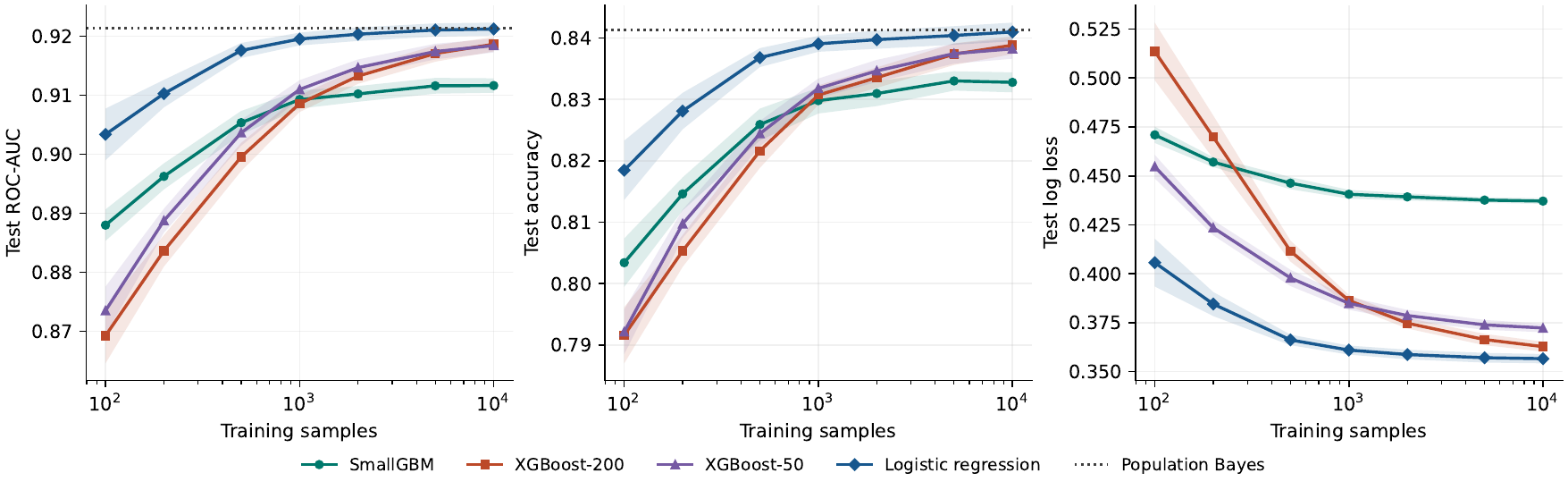}
    \caption{Generalization on the Gaussian mixture. The panels report test ROC-AUC, accuracy and log loss, respectively. Points are means over 20 seeds, with pointwise 95\% $t$ intervals. Dotted lines show population Bayes ROC-AUC and accuracy. The linear control had the highest mean ROC-AUC at all seven sizes. Both XGBoost variants first exceeded SmallGBM in mean accuracy at $n=1000$.}
    \label{fig:n_crossover}
\end{figure}

At $n=100$, mean test ROC-AUC was 0.9034 for logistic regression, 0.8880 for SmallGBM, 0.8735 for \texttt{XGBoost-50} and 0.8692 for \texttt{XGBoost-200}. The paired mean logistic-minus-SmallGBM difference was $+0.0154$ (pointwise 95\% $t$ interval $[0.0116,0.0191]$). At $n=10{,}000$, the corresponding scores were 0.9212, 0.9116, 0.9184 and 0.9186. The linear control approaches the Bayes ROC-AUC for this data-generating process.

We computed the XGBoost-minus-SmallGBM test ROC-AUC difference within each seed. For \texttt{XGBoost-50}, the mean difference changed from $-0.0017$ at $n=500$ to $+0.0017$ at $n=1000$. For \texttt{XGBoost-200}, it changed from $-0.0007$ at $n=1000$ to $+0.0030$ at $n=2000$. These pairs of measured sizes bracket the observed reversals in mean ROC-AUC for this distribution. The grid leaves the ordering at intermediate sizes unresolved.

\subsection{Surviving Data Pathologies}
\label{sec:pathologies}

\subsubsection{Heavy-Tailed Contamination and Label Corruption}
\label{sec:heavy_tailed_noise}

We varied the proportion of corrupted training targets over $0,1,5,10$ and $20\%$ in ten paired trials ($n_{\rm train}=300$, $d=10$). Each task used a fixed clean test set of 5,000 observations. For classification, $X\sim\mathcal N(0,I_d)$ and $Y=\mathbf{1}\{1.5X_1-X_2+0.8X_3X_4+0.5\epsilon>0\}$, where $\epsilon\sim\mathcal N(0,1)$. Randomly selected binary labels were flipped. For Friedman~\#1 regression with unit Gaussian noise, we added independent standard Cauchy draws scaled by the interquartile range of clean training responses to the selected targets. No draws were clipped. All XGBoost models in this sweep used 200 trees, depth 3 and learning rate 0.05. The classifier used logistic loss. Regression comparators used squared error or absolute error. The median constant was fitted to the available training targets.

\begin{figure}[H]
    \centering
    \includegraphics[width=0.98\textwidth]{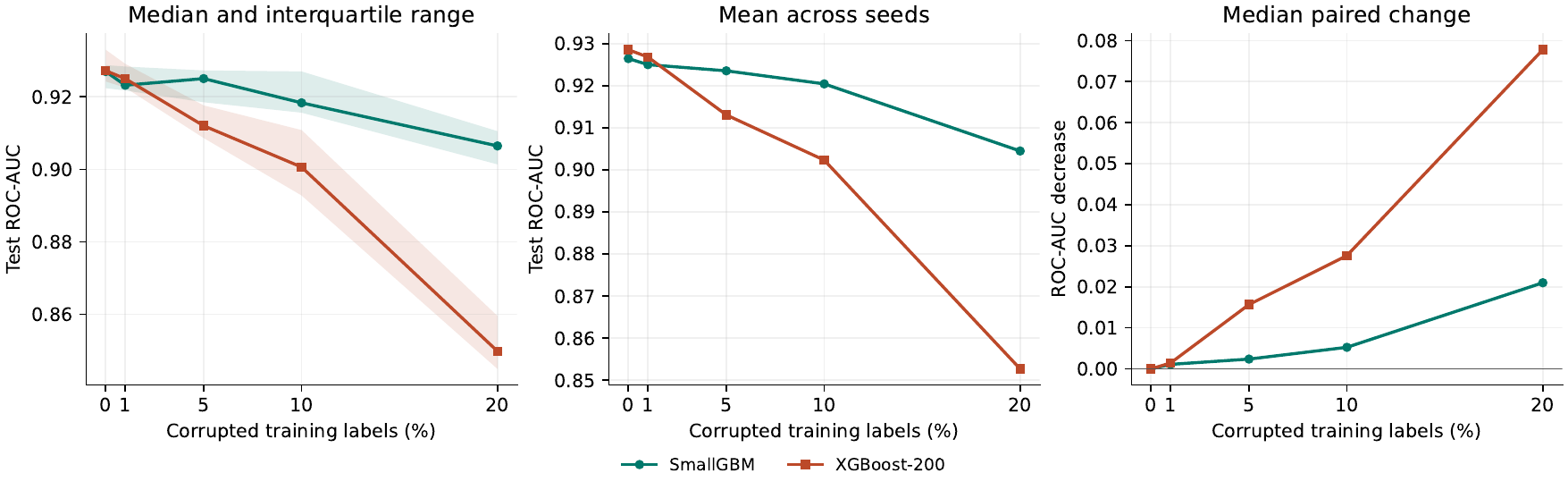}
    \caption{Test ROC-AUC under binary training-label flips. Left to right: median ROC-AUC with interquartile range, mean ROC-AUC and median within-trial AUC loss relative to the clean fit. At 20\% flips, the median paired losses were 0.0210 for SmallGBM and 0.0778 for the fixed XGBoost-200 classifier.}
    \label{fig:label_noise_cls}
\end{figure}

Mean ROC-AUC on clean labels was 0.9265 for SmallGBM and 0.9286 for XGBoost-200. At 20\% flips it fell to 0.9045 and 0.8527, respectively. For each seed we subtracted its corrupted-data AUC from its clean-data AUC, then took each model's median across the ten seeds. The XGBoost-200 median loss was 3.7 times the SmallGBM loss, and SmallGBM lost less AUC in all ten trials. The comparison changes the entire tree implementation. A within-SmallGBM leaf-estimator ablation would be needed to attribute that difference to the median alone.

\begin{figure}[H]
    \centering
    \includegraphics[width=0.98\textwidth]{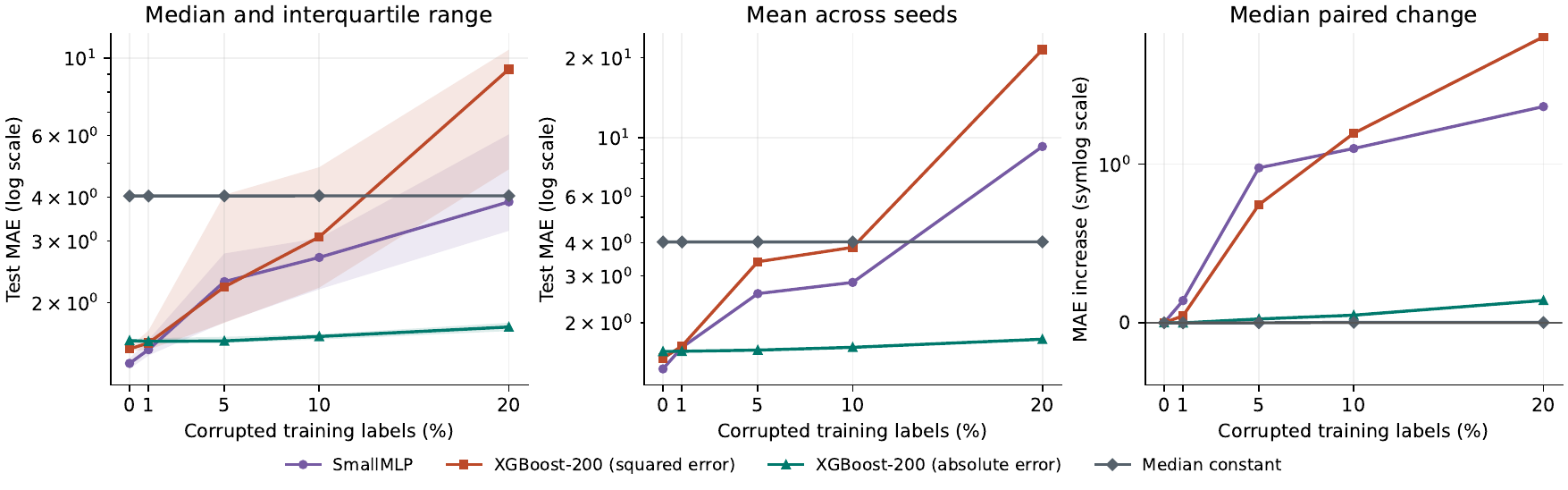}
    \caption{Friedman~\#1 regression under additive Cauchy contamination. Left to right: median test MAE with interquartile range, mean test MAE and median within-trial MAE increase from the clean fit, across ten trials. The two XGBoost-200 configurations differ in their training loss.}
    \label{fig:cauchy_noise_reg}
\end{figure}

\begin{table}[H]
\centering
\caption{Mean and Median Test Performance Across Progressive Target Corruption (10/10 Seeds Retained)}
\vspace{4pt}
\resizebox{\textwidth}{!}{%
\begin{tabular}{llccccc}
\toprule
\textbf{Task / Metric} & \textbf{Model} & \textbf{0\% (Clean)} & \textbf{1\% Noise} & \textbf{5\% Noise} & \textbf{10\% Noise} & \textbf{20\% Noise} \\
\midrule
Classification (ROC-AUC $\uparrow$) & \textbf{SmallGBM} & $0.9265$ ($0.9270$) & $0.9250$ ($0.9232$) & $\mathbf{0.9236}$ ($\mathbf{0.9250}$) & $\mathbf{0.9204}$ ($\mathbf{0.9183}$) & $\mathbf{0.9045}$ ($\mathbf{0.9065}$) \\
 & XGBoost-200 (fixed) & $\mathbf{0.9286}$ ($\mathbf{0.9272}$) & $\mathbf{0.9268}$ ($\mathbf{0.9250}$) & $0.9131$ ($0.9120$) & $0.9023$ ($0.9006$) & $0.8527$ ($0.8499$) \\
\midrule
Regression (Test MAE $\downarrow$) & \textbf{SmallMLP} & $\mathbf{1.338}$ ($\mathbf{1.339}$) & $1.613$ ($\mathbf{1.466}$) & $2.576$ ($2.294$) & $2.839$ ($2.691$) & $9.260$ ($3.882$) \\
 & XGBoost-200 ($L_2$) & $1.468$ ($1.473$) & $1.632$ ($1.532$) & $3.399$ ($2.216$) & $3.852$ ($3.080$) & $21.494$ ($9.301$) \\
 & XGBoost-200 ($L_1$) & $1.557$ ($1.556$) & $\mathbf{1.559}$ ($1.548$) & $\mathbf{1.576}$ ($\mathbf{1.553}$) & $\mathbf{1.614}$ ($\mathbf{1.599}$) & $\mathbf{1.732}$ ($\mathbf{1.702}$) \\
 & Median Constant & $4.036$ ($4.033$) & $4.036$ ($4.033$) & $4.038$ ($4.035$) & $4.039$ ($4.038$) & $4.039$ ($4.038$) \\
\bottomrule
\multicolumn{7}{l}{\footnotesize \textit{Note:} Values report $\text{Mean} \; (\text{Median})$ across 10 random seeds.}
\end{tabular}%
}
\label{tab:noise_sweep}
\end{table}

On clean data SmallMLP had the lowest mean MAE (1.338). At 20\% Cauchy contamination, its mean MAE reached 9.260 and its median MAE was 3.882. The squared-error XGBoost-200 fit gave a mean of 21.494 and a median of 9.301. A single seed contributed errors of 54.57 and 145.43, respectively. Under the same contamination, the absolute-error XGBoost-200 configuration had mean MAE 1.732, compared with 1.557 on clean data. These finite-trial summaries illustrate sensitivity to the training objective and to extreme draws. They do not identify the effect of a component of the neural network.

\subsubsection{Prediction Interval Coverage Under Heteroscedasticity}
\label{sec:conformal_uq}

We evaluate the test coverage and width of the prediction intervals described in Section~\ref{sec:uq} on a heteroscedastic nonlinear regression problem ($n = 150$, $n_{\text{test}} = 2000$, 20 independent random seeds, target coverage $1 - \alpha = 0.90$). Covariates are drawn as $X \sim \mathcal{N}(0, I_d)$ across both a low-dimensional regime ($d = 10$) and a high-dimensional regime ($d = 100$, where $d \approx n$), with targets generated via $Y = 1.5 X_1 - X_2 + 0.8 X_3 X_4 + \sin(X_5) + \sigma(X)\varepsilon$, $\varepsilon \sim \mathcal{N}(0, 1)$, where $\sigma(X) = 0.25$ for $X_1 \le 0$ and $\sigma(X) = 1.50$ for $X_1 > 0$ (a $6\times$ noise-scale shift). We compare a parametric Gaussian Process (\texttt{GP (Type-II ML)}, ARD-RBF + WhiteKernel fitted on all $n = 150$ samples) against \texttt{SmallMLP} calibrated via a single non-tuned $75/25$ split ($n_{\text{fit}} = 112$, $n_{\text{cal}} = 38$) under both unweighted split conformal prediction (Equation~\eqref{eq:conformal_cov}) and locally weighted conformal prediction over the penultimate bottleneck embedding $h(x)$ (Equation~\eqref{eq:h_cal}).

\begin{figure}[H]
    \centering
    \includegraphics[width=0.98\textwidth]{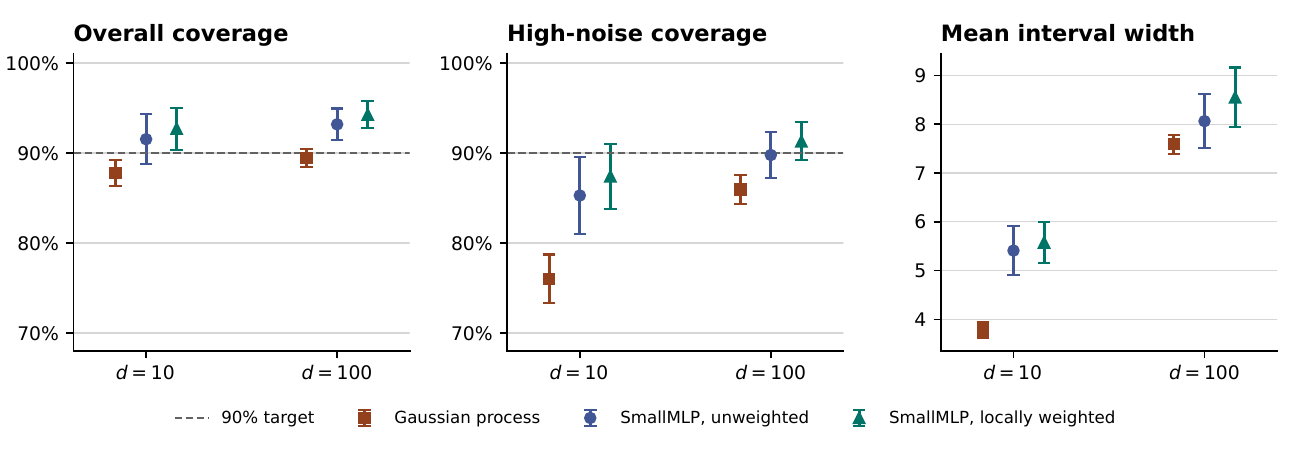}
    \caption{Prediction intervals with a sixfold difference in noise scale. Panels show empirical marginal coverage, coverage in the high-noise subgroup and mean interval width. Points are means over the same 20 seeds per dimension, with pointwise 95\% $t$ intervals. The coverage target is 90\%.}
    \label{fig:conformal_uq}
\end{figure}

With $m=38$ and $1-\alpha=0.90$, ordinary split conformal uses $k=\lceil39\times0.90\rceil=36$. We directly select the 36th ordered calibration residual. With almost surely distinct exchangeable scores its marginal coverage over calibration sets and future observations is $36/39\approx92.3\%$. Realized benchmark averages can differ.

At $d=10$, the ordinary split interval covered $91.5\%$ of test responses with a mean width of 5.410. Local weighting raised coverage to $92.7\%$ and width to 5.575. The paired change in width was $+0.165$ (pointwise 95\% $t$ interval $[-0.023,0.353]$). At $d=100$, the corresponding changes were from $93.2\%$ to $94.3\%$ in coverage and from 8.064 to 8.550 in width. The width increased in all 20 paired trials (mean difference $+0.487$, interval $[0.137,0.836]$). The comparisons use the same fitted network and calibration residuals within each trial. The two rules attain different observed coverages, so these results do not measure width efficiency at exactly matched coverage.

In the high-noise subgroup at $d=10$, observed coverage was $85.3\%$ for ordinary split conformal, $87.4\%$ for local weighting and $76.0\%$ for the fitted Gaussian process. At $d=100$, the corresponding estimates were $89.8\%$, $91.3\%$ and $86.0\%$. The weighted interval's mean widths in the low- and high-noise groups were 5.39 and 5.75 at $d=10$, compared with 8.57 and 8.53 at $d=100$. Thus the difference in width across noise groups largely disappeared in the higher-dimensional experiment. The finite-radius cap described in Section~\ref{sec:uq} was used for $1.14\%$ and $1.22\%$ of test predictions, respectively.

\subsection{Component Ablations and Meta-Validation}

\subsubsection{Sensitivity of Classification Accuracy to Width Coefficients}
We repeated the 45-point classification width grid on three development datasets specified before the rerun: Iris (150 observations), Wine (178) and Statlog Heart (270). Each setting used the same five stratified folds and random seed 42. The median imputer was fitted on the training observations in each fold. Accuracy was averaged first across folds and then across the three datasets. This sensitivity check uses a narrower panel than the historical development grid. The regression backbone uses equal-width layers and is independent of these coefficients.

Mean accuracy ranged from 0.9146 to 0.9288, with the highest cell at $(\alpha,\beta)=(7,0.6)$. The coefficients already in use, $(6,0.7)$, gave 0.9233 (0.9533 on Iris, 0.9833 on Wine and 0.8333 on Statlog Heart). They were 0.55 percentage points below the highest cell. This grid describes sensitivity with one split and initialization seed. The three development datasets are distinct from the six retrospective evaluation datasets in Table~\ref{tab:real_data}.

\subsubsection{Depth Ablation at Fixed Training Settings}
\label{sec:depth_ablation}

We varied the hidden-layer count $L$ from two to five on Friedman~\#1 regression and nonlinear binary classification, using $n\in\{100,200,400\}$, $d=10$, ten paired seeds and 5,000 independent test observations per trial. Width rules and optimization settings were fixed across depths (maximum 500 epochs, early-stopping patience 30). Parameter count increased with depth.

\begin{figure}[H]
    \centering
    \includegraphics[width=0.96\textwidth]{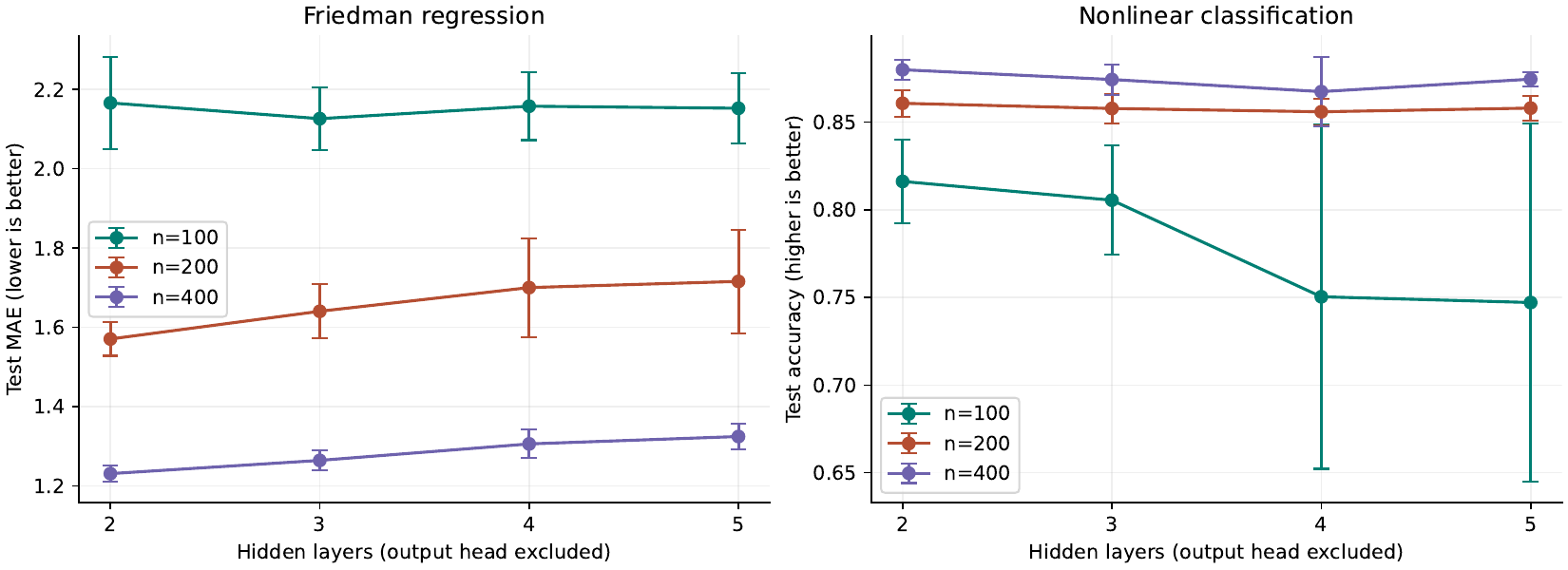}
    \caption{Test MAE for Friedman~\#1 regression and test accuracy for nonlinear classification across depths and training sizes. Points show ten-trial means with pointwise 95\% $t$ intervals. Paired comparisons in the text use the same seed at both depths.}
    \label{fig:depth_ablation}
\end{figure}

At $n=100$, the two-layer classifier exceeded the five-layer variant in mean accuracy by 6.9 percentage points, with a wide paired 95\% $t$ interval from $-1.8$ to $+15.6$ points. For Friedman regression at $n=400$, adding three layers increased MAE by 0.094 (paired interval $[0.065,0.122]$). These observations support using two layers as a practical default on the tasks examined. The ablation changes both depth and parameter count.

\subsection{Gaussian Location Estimation}
\label{sec:negative_results}

We examine the efficiency of estimating a Gaussian location and relate this separate experiment to the observed classification results in Figure~\ref{fig:n_crossover}.

\subsubsection{Sample Mean and Sample Median}
\label{sec:gaussian_penalty}

We drew independent samples from $\mathcal{N}(0,1)$ and used the sample mean and the sample median to estimate the common center at zero. At each sample size we averaged squared estimation errors over 5,000 draws.

\begin{figure}[H]
    \centering
    \includegraphics[width=0.96\textwidth]{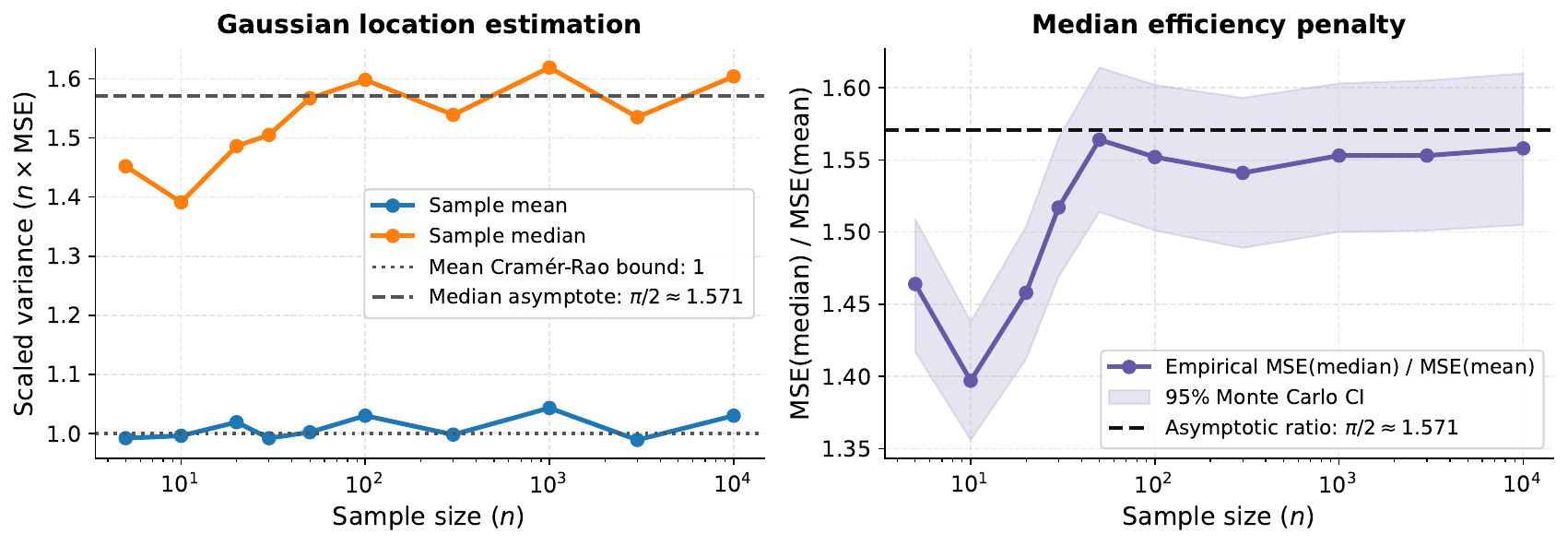}
    \caption{\textbf{The Gaussian Efficiency Penalty of Robust Estimation.} Left: On uncorrupted Gaussian data, the sample mean achieves the Cram\'{e}r-Rao lower bound ($n \times \text{MSE} \to 1$), whereas $n$ times the MSE of the sample median converges to $\pi/2 \approx 1.571$. Right: The empirical MSE ratio $\text{MSE}(\text{median}) / \text{MSE}(\text{mean})$ across sample sizes, reaching $1.558$ ($95\%$ Monte Carlo CI: $[1.505, 1.610]$) at $n = 10^4$. Both panels concern estimation of a Gaussian location.}
    \label{fig:gaussian_penalty}
\end{figure}

As shown in Figure~\ref{fig:gaussian_penalty}, while the sample mean is the minimum-variance unbiased estimator under Gaussian noise ($n \times \text{MSE} \approx 1.00$ across all $n$), the sample median incurs a strict variance penalty governed by the asymptotic distribution of sample quantiles:
\begin{equation}
\label{eq:asymptotic_efficiency}
\lim_{n\to\infty} \frac{\operatorname{Var}\left(\operatorname{median}(X_1,\ldots,X_n)\right)}{\operatorname{Var}(\bar X_n)} = \frac{1}{4 f(0)^2 \sigma^2} = \frac{\pi}{2} \approx 1.5708
\end{equation}
where $f(0)$ is the density of $\mathcal{N}(0,\sigma^2)$ at its median. The resulting asymptotic relative efficiency of the median is $2/\pi$. This calculation describes an independent Gaussian sample. At each boosting step the classification pseudo-residuals depend on binary labels and the current fitted probabilities. Attributing the ROC-AUC reversals in Figure~\ref{fig:n_crossover} to the leaf estimator would require varying that component while holding the other tree settings fixed.

The choice of metric also affects the comparison. At $n=100$, \texttt{XGBoost-50} had lower mean test log loss than SmallGBM (0.455 versus 0.471), even though SmallGBM ranked the classes better by ROC-AUC. At $n=10{,}000$, the corresponding log losses were 0.372 and 0.437. Log loss measures the quality of probability predictions. Identifying systematic underconfidence would require a separate calibration analysis.
\subsection{Performance on Six Public Datasets}\label{sec:real_results}
We evaluated three fixed configurations on each dataset using five outer folds repeated with seeds 11, 29 and 47. Within each repeat, all three models used the same folds. The linear baseline was standardized logistic regression ($C=1$) for classification or ridge regression (penalty 1) for regression. The tree baseline used XGBoost's library defaults. The third model was SmallGBM for classification and SmallMLP for regression. Imputation, encoding and scaling used the outer training fold. For Energy we predicted heating load and excluded cooling load.

Classification splits were stratified over the splitting units. Duplicate Ionosphere predictor vectors shared a fold. Parkinson's recordings were grouped by subject identifier, with each subject's held-out probabilities averaged before computing ROC-AUC. The released file contains 32 distinct identifier prefixes, although the UCI description reports 31 participants. For Yacht and Energy we held out entire hull or building geometries, respectively. For each repeat we pooled out-of-fold predictions to obtain one metric. Table~\ref{tab:real_data} reports the arithmetic mean across repeats. The repeats reuse observations, and the few independent subjects or designs limit the precision of these estimates.

\begin{table}[H]
\centering
\caption{Performance on public datasets. Dataset names link to the UCI source. The last column uses the SmallGBM classifier or SmallMLP regressor. Bold marks the best observed mean within each row.}
\label{tab:real_data}
\small
\begin{tabular}{@{}lrrlrrr@{}}
\toprule
Dataset & Rows & Features & Split units & Linear & XGBoost & Small model\\
\midrule
\multicolumn{7}{@{}l}{\textit{Classification, ROC-AUC (higher is better)}}\\
\href{https://doi.org/10.24432/C5Q59J}{Hepatitis} & 155 & 19 & 155 rows & \textbf{0.850} & 0.822 & 0.758\\
\href{https://doi.org/10.24432/C5DW2B}{Breast Cancer (WDBC)} & 569 & 30 & 569 rows & \textbf{0.995} & 0.993 & 0.987\\
\href{https://doi.org/10.24432/C5W01B}{Ionosphere} & 351 & 34 & 350 vectors & 0.896 & \textbf{0.967} & 0.931\\
\href{https://doi.org/10.24432/C59C74}{Parkinson's} & 195 & 22 & 32 subject IDs & 0.734 & 0.757 & \textbf{0.778}\\
\addlinespace
\multicolumn{7}{@{}l}{\textit{Regression, MAE in each target's units (lower is better)}}\\
\href{https://doi.org/10.24432/C5XG7R}{Yacht Hydrodynamics} & 308 & 6 & 22 hulls & 7.148 & \textbf{0.571} & 1.459\\
\href{https://doi.org/10.24432/C51307}{Energy Efficiency} & 768 & 8 & 12 geometries & \textbf{3.735} & 6.325 & 5.513\\
\bottomrule
\end{tabular}
\end{table}

The linear model had the best observed mean on Hepatitis, Breast Cancer and Energy. XGBoost led on Ionosphere and Yacht, and SmallGBM led on Parkinson's. The point differences on Breast Cancer were small. These are retrospective benchmarks. Earlier SmallMLP development scripts included the six datasets, with Breast Cancer also used in an architecture ablation. An independently collected cohort would provide a stronger test of transfer.

\section{Conclusion}
\label{sec:conclusion}

A gain measured while selecting hyperparameters can overstate the gain after refitting. At $n=100$, cross-validation on random labels overstated the gain from tuning by 0.063 ROC-AUC, while tuning on a Gaussian task still improved held-out performance. On the separate Gaussian scaling benchmark, logistic regression had the highest mean ROC-AUC across the grid. Selecting a model whose decision boundary matches the signal can matter even at the smallest training size.

SmallGBM lost less discrimination than the tested XGBoost-200 configuration under random binary label flips. Heavy-tailed regression responses produced a different ordering, with the absolute-error boosting configuration attaining the lowest mean error at 20\% contamination. Under heteroscedastic noise, local interval weights increased both average coverage and average width relative to ordinary split conformal. Whole-model comparisons, repeated trials of a single generator and public-data folds each support different conclusions. Extending these results to new applications requires an evaluation on the relevant outcomes and sampling units.

\section*{Impact Statement}
A small laboratory or clinical research group may have enough observations to fit several predictors but too few to estimate small performance differences precisely. On random labels, selecting the highest cross-validation score produced an apparent improvement even though the features contained no signal. Recording the search and evaluating the selected model on independent observations can prevent an optimistic estimate from guiding which measurements or experiments receive resources. The Gaussian experiment also illustrates the value of retaining a low-cost linear comparator when the signal may be simple.

Prediction intervals affect which cases receive additional measurements or review. In our simulation at $d=10$, the locally weighted method covered about 93\% of responses overall and 87\% in the high-noise group. A team planning around the overall figure would underestimate missed responses in that group. Reporting group-specific coverage together with interval width makes that trade-off visible. The clinical benchmarks measure predictions on retrospective public data. Discrimination on these data does not establish clinical benefit, and a patient-level split cannot measure transport to a new hospital. Decisions affecting patients would require evaluation in the intended population, including the consequences of missed cases and unnecessary follow-up.

\section*{Data and Code Availability}
The accompanying source package includes the revised benchmark scripts, dataset identifiers, split protocols and numerical summaries. The project repository is \url{https://github.com/nsdmlk/SmallData-Trilemma}. Public dataset sources and the units used for splitting are also identified in Section~\ref{sec:real_results}.

\section*{Acknowledgements}

The authors thank Beijing Institute of Technology for providing
a supportive academic environment and opportunities for research
and collaboration.

\section*{Author Contributions}
Ilya Emelyanov conceived the study, developed the mathematical framework, and implemented the SmallGBM and SmallMLP libraries. Dmitrii Dobrydin designed and conducted the benchmarks and experiments. Muhammad Hamza contributed to validation, helped revise the manuscript structure, and provided methodological feedback. Malak Abid Ali Khan provided research supervision, critical review of the methodology, and manuscript editing.

\bibliographystyle{unsrt}
\bibliography{references}

@inproceedings{chen2016xgboost,
  title={XGBoost: A Scalable Tree Boosting System},
  author={Chen, Tianqi and Guestrin, Carlos},
  booktitle={Proceedings of the 22nd ACM SIGKDD International Conference on Knowledge Discovery and Data Mining},
  pages={785--794},
  year={2016},
  organization={ACM}
}

@inproceedings{ke2017lightgbm,
  title={LightGBM: A Highly Efficient Gradient Boosting Decision Tree},
  author={Ke, Guolin and Meng, Qi and Finley, Thomas and Wang, Taifeng and Chen, Wei and Ma, Weidong and Ye, Qiwei and Liu, Tie-Yan},
  booktitle={Advances in Neural Information Processing Systems (NeurIPS)},
  volume={30},
  pages={3146--3154},
  year={2017}
}

@book{vovk2005algorithmic,
  title={Algorithmic learning in a random world},
  author={Vovk, Vladimir and Gammerman, Alexander and Shafer, Glenn},
  year={2005},
  publisher={Springer Science \& Business Media}
}

@article{angelopoulos2023gentle,
  title={A gentle introduction to conformal prediction and distribution-free uncertainty quantification},
  author={Angelopoulos, Anastasios N and Bates, Stephen},
  journal={Foundations and Trends in Machine Learning},
  volume={16},
  number={4},
  pages={494--591},
  year={2023},
  publisher={Now Publishers, Inc.}
}

@book{huber2009robust,
  title={Robust statistics},
  author={Huber, Peter J and Ronchetti, Elvezio M},
  year={2009},
  publisher={John Wiley \& Sons}
}

@article{guan2023lcp,
  author  = {Guan, Leying},
  title   = {Localized conformal prediction: a generalized inference framework for conformal prediction},
  journal = {Biometrika},
  year    = {2023},
  volume  = {110},
  number  = {1},
  pages   = {33--50},
  doi     = {10.1093/biomet/asac040}
}

@article{hore2025rlcp,
  author  = {Rohan Hore and Rina Foygel Barber},
  title   = {Conformal prediction with local weights: randomization enables robust guarantees},
  journal = {Journal of the Royal Statistical Society Series B: Statistical Methodology},
  year    = {2025},
  volume  = {87},
  number  = {2},
  pages   = {549--578},
  doi     = {10.1093/jrsssb/qkae103}
}

@article{cawley2010selection,
  author={Gavin C. Cawley and Nicola L. C. Talbot},
  title={On Over-fitting in Model Selection and Subsequent Selection Bias in Performance Evaluation},
  journal={Journal of Machine Learning Research}, year={2010}, volume={11}, pages={2079--2107}
}

@article{varoquaux2018cross,
  author={Gaël Varoquaux}, title={Cross-validation failure: Small sample sizes lead to large error bars},
  journal={NeuroImage}, year={2018}, volume={180}, pages={68--77}
}

@inproceedings{kadra2021well,
  author={Arlind Kadra and Marius Lindauer and Frank Hutter and Josif Grabocka},
  title={Well-tuned Simple Nets Excel on Tabular Datasets},
  booktitle={Advances in Neural Information Processing Systems}, year={2021}, volume={34}, pages={23928--23941}
}

@inproceedings{grinsztajn2022tree,
  author={Léo Grinsztajn and Édouard Oyallon and Gaël Varoquaux},
  title={Why do tree-based models still outperform deep learning on typical tabular data?},
  booktitle={Advances in Neural Information Processing Systems}, year={2022}, volume={35}, pages={507--520}
}

@article{hollmann2025tabpfn,
  author={Noah Hollmann and Samuel Müller and Lennart Purucker and Arjun Krishnakumar and Max Körfer and Shi Bin Hoo and Robin Tibor Schirrmeister and Frank Hutter},
  title={Accurate predictions on small data with a tabular foundation model},
  journal={Nature}, year={2025}, volume={637}, pages={319--326}, doi={10.1038/s41586-024-08328-6}
}

@inproceedings{arik2021tabnet,
  author = {Sercan {\"O}. Arik and Tomas Pfister},
  title = {{TabNet}: Attentive Interpretable Tabular Learning},
  booktitle = {Proceedings of the AAAI Conference on Artificial Intelligence},
  volume = {35}, pages = {6679--6687}, year = {2021},
  doi = {10.1609/aaai.v35i8.16826}
}

@inproceedings{gorishniy2021revisiting,
  author = {Yury Gorishniy and Ivan Rubachev and Valentin Khrulkov and Artem Babenko},
  title = {Revisiting Deep Learning Models for Tabular Data},
  booktitle = {Advances in Neural Information Processing Systems},
  volume = {34}, pages = {18932--18943}, year = {2021},
  url = {https://proceedings.neurips.cc/paper/2021/hash/9d86d83f925f2149e9edb0ac3b49229c-Abstract.html}
}

@inproceedings{feurer2015automl,
  author = {Matthias Feurer and Aaron Klein and Katharina Eggensperger and Jost Springenberg and Manuel Blum and Frank Hutter},
  title = {Efficient and Robust Automated Machine Learning},
  booktitle = {Advances in Neural Information Processing Systems},
  volume = {28}, year = {2015},
  url = {https://proceedings.neurips.cc/paper/2015/hash/11d0e6287202fced83f79975ec59a3a6-Abstract.html}
}

@article{erickson2020autogluon,
  author = {Nick Erickson and Jonas Mueller and Alexander Shirkov and Hang Zhang and Pedro Larroy and Mu Li and Alexander Smola},
  title = {{AutoGluon-Tabular}: Robust and Accurate {AutoML} for Structured Data},
  journal = {arXiv preprint arXiv:2003.06505}, year = {2020},
  url = {https://arxiv.org/abs/2003.06505}
}

@book{rasmussen2006gpml,
  author = {Carl Edward Rasmussen and Christopher K. I. Williams},
  title = {Gaussian Processes for Machine Learning},
  publisher = {MIT Press}, year = {2006},
  isbn = {0-262-18253-X}, url = {https://gaussianprocess.org/gpml/}
}

@inproceedings{lakshminarayanan2017ensembles,
  author = {Balaji Lakshminarayanan and Alexander Pritzel and Charles Blundell},
  title = {Simple and Scalable Predictive Uncertainty Estimation using Deep Ensembles},
  booktitle = {Advances in Neural Information Processing Systems},
  volume = {30}, year = {2017},
  url = {https://proceedings.neurips.cc/paper/2017/hash/9ef2ed4b7fd2c810847ffa5fa85bce38-Abstract.html}
}

@inproceedings{romano2019cqr,
  author = {Yaniv Romano and Evan Patterson and Emmanuel Cand{\`e}s},
  title = {Conformalized Quantile Regression},
  booktitle = {Advances in Neural Information Processing Systems},
  volume = {32}, year = {2019},
  url = {https://proceedings.neurips.cc/paper/2019/hash/5103c3584b063c431bd1268e9b5e76fb-Abstract.html}
}

@article{ramprasad2017materials,
  author = {Rampi Ramprasad and Rohit Batra and Ghanshyam Pilania and Arun Mannodi-Kanakkithodi and Chiho Kim},
  title = {Machine learning in materials informatics: recent applications and prospects},
  journal = {npj Computational Materials}, volume = {3}, pages = {54}, year = {2017},
  doi = {10.1038/s41524-017-0056-5}
}

@article{radovic2018physics,
  author = {Alexander Radovic and Mike Williams and David Rousseau and Michael Kagan and Daniele Bonacorsi and Alexander Himmel and Adam Aurisano and Kazuhiro Terao and Taritree Wongjirad},
  title = {Machine learning at the energy and intensity frontiers of particle physics},
  journal = {Nature}, volume = {560}, pages = {41--48}, year = {2018},
  doi = {10.1038/s41586-018-0361-2}
}
\end{document}